\documentclass{ifacconf}

\usepackage{amsmath}
\usepackage{commath}
\usepackage{tikz}
\usepackage{graphicx}      % include this line if your document contains figures
\usepackage{pgfplots}
\pgfplotsset{compat=newest}
\pgfplotsset{plot coordinates/math parser=false}
\newlength\figureheight
\newlength\figurewidth

\usetikzlibrary{bayesnet}

\usepackage{graphicx}      % include this line if your document contains figures
\usepackage{natbib}        % required for bibliography

\begin{document}
\begin{frontmatter}

\title{Multi-agent Gaussian Process Motion Planning via Probabilistic Inference\thanksref{footnoteinfo}}

\thanks[footnoteinfo]{This research has been supported by the European Regional Development Fund under the grant KK.01.1.1.01.0009 (DATACROSS).}

\author[First]{Luka Petrovi\' c, Ivan Markovi\' c, Marija Seder}
\address[First]{University of Zagreb Faculty of Electrical Engineering and
Computing, Laboratory for Autonomous Systems and Mobile Robotics, Croatia (e-mail: \{luka.petrovic, ivan.markovic, marija.seder\}@fer.hr)}

\begin{abstract}                % Abstract of not more than 250 words.
This paper deals with motion planning for multiple agents by representing the problem as a simultaneous optimization of every agent's trajectory. Each trajectory is considered as a sample from a one-dimensional continuous-time Gaussian process (GP) generated by a linear time-varying stochastic differential equation driven by white noise. By formulating the planning problem as probabilistic inference on a factor graph, the structure of the pertaining GP can be exploited to find the solution efficiently using numerical optimization. In contrast to planning each agent's trajectory individually, where only the current poses of other agents are taken into account, we propose simultaneous planning of multiple trajectories that works in a predictive manner. It takes into account the information about each agent's whereabouts at every future time instant, since full trajectories of each agent are found jointly during a single optimization procedure. We compare the proposed method to an individual trajectory planning approach, demonstrating significant improvement in both success rate and computational efficiency.
\end{abstract}

\begin{keyword}
motion planning, trajectory optimization, multi-agent system, probabilistic inference, factor graphs
\end{keyword}

\end{frontmatter}
%===============================================================================

\section{Introduction}
Motion planning is an indispensable skill for robots that aspire to navigate through an environment without collisions.
Motion planning algorithms attempt to generate trajectories through the robot's configuration space that are both
feasible and optimal based on some performance criterion that may vary depending on the task, robot, or environment. Motion planning algorithms that can be executed in real time are highly encouraged, mostly
because they allow for fast replanning in response to environment changes.

A significant amount of recent work has focused on trajectory optimization and related problems. Trajectory optimization methods start with an initial trajectory and then minimize an objective function in order to optimize the trajectory. \cite{chomp} and \cite{chomp-ijrr}, in their work abbreviated as CHOMP, proposed utilizing a precomputed signed distance field for fast collision checking and using covariant gradient descent to minimize obstacle and smoothness costs. \cite{stomp} developed a stochastic trajectory optimization method (STOMP) that samples a series of noisy trajectories to explore the space around an initial trajectory which are then combined to produce an updated trajectory with lower cost. The key trait of STOMP is its ability to optimize non-differentiable constraints. An important shortcoming of CHOMP and STOMP is the need for many trajectory states in order to reason about fine resolution obstacle representations or find feasible solutions when there are many constraints.  \cite{trajopt}, \cite{schulman2014motion}, in their framwork called TrajOpt, formulate motion planning as sequential quadratic programming. The key feature of TrajOpt is the ability to solve complex motion planning problems with few states since swept volumes are considered to ensure continuous-time safety. However, if the smoothness is required in the output trajectory, either a densely parametrized trajectory or post-processing of the trajectory might still be needed thus increasing computation time. In order to overcome the computational cost incurred by using large number of states, the Gaussian process motion planning family of algorithms (\cite{gpmp}, \cite{gpmp2}, \cite{gpmpgraph}) use continous-time trajectory representation. \cite{gpmp} parametrize the trajectory with a few support states and then use Gaussian process (GP) interpolation to query the trajectory at any time of interest. \cite{gpmp2}, in their framework called GPMP2, represent continuous-time trajectories as samples from a GP and then formulate the planning problem as probabilistic inference, generating fast solutions by exploiting the sparsity of the underlying linear system. A useful property of GPMP2 is its extensibility and applicability for wide range of problems (\cite{rana2017towards}, \cite{steap}, \cite{maric2018singularity}) and in this paper we also rely on the aforementioned framework. All mentioned methods are primarily designed for planning motion of one robot in static environments. In many challenging problems, such as warehouse management, delivery and construction, a single robot may not be able to achieve desired tasks and therefore multi-robot teams are required. Multi-robot teams are also more robust to malfunctions, since one robot can take over the tasks of another robot in case of failure.

In this paper, we propose using a continous time Gaussian process trajectory representations in order to plan motion for every robot in a multi-agent system. We augment the method proposed in \cite{gpmp2} and consider multi-agent trajectory optimization from a probabilistic inference perspective, optimizing all of the trajectories concurrently. By optimizing the trajectories at the same time, our method takes into account the information where each agent will be at every future time instant, thus working in a predictive manner. We evaluated our approach in simulation and compared it to planning each agent's trajectory indvidually.

The rest of the paper is organized as follows. Section \ref{subsec:gaussian_trajectory} presents Gaussian processes as trajectory representations. In Section \ref{subsec:multiag}, the method for simultaneous multi-agent trajectory optimization as probabilistic inference is proposed. Section \ref{subsec:impdet} describes implementation aspects of the proposed method. Section \ref{subsec:results} demonstrates the main results in simulation and Section \ref{subsec:conclusion} concludes the paper.

\section{Gaussian processes as trajectory representations}\label{subsec:gaussian_trajectory}

A continuous-time trajectory is considered as a sample from a vector-valued Gaussian process (GP), $x(t) \sim \mathcal{GP}(\mu(t), K(t, t^{\prime})) $, with mean $\mu(t)$ and covariance $ K(t, t^{\prime})$, generated by a linear time-varying stochastic differential equation (LTV-SDE)
\begin{equation}
\dot x(t) = A(t) x(t) + u(t) + F(t) w(t)
\label{ltvsde}
\end{equation}
where $A$, $F$ are system matrices, $u$ is a known control input and $w(t)$ is generated by a white noise process. The white noise process is itself a GP with zero mean value
\begin{equation}
w(t) \sim \mathcal{GP} (0, Q_c \delta (t - t^{\prime})),
\end{equation}
where $Q_c$ is a power spectral density matrix .

The solution of the LTV-SDE in (\ref{ltvsde}) is generated by the mean and covariance of the GP:
\begin{equation}
\mu (t) = \Phi(t, t_0) \mu_0 + \int_{t_0}^t \Phi (t, s) u(s) ds
\end{equation}
\begin{multline}
K(t, t^{\prime}) =  \Phi(t, t_0) K_{0} \Phi(t^{\prime}, t_0)^T + \\ \int_{t_0}^{min(t,t^{\prime})} \Phi(t, s)  F(s) Q_{c} F(s)^T \Phi(t^{\prime}, s) ds,
\end{multline}
where $\mu_0$, $K_0$ are initial mean and covariance of the first state, and $\Phi (t, s)$ is the state transition matrix (\cite{barfoot2014batch}).

The GP prior distribution is then given in terms of its mean  $\mu$ and covariance $K$:
\begin{equation}
p(x) \propto \exp  \{  -\frac{1}{2} \norm{x - \mu }_K^2   \}.
\label{eq:prior}
\end{equation}

One major benefit of using Gaussian processes to model continuous-time trajectory in motion planning is the possibility to query the planned state $x(\tau)$ at any time of interest $\tau$, and not only at discrete time instants. If the prior proposed in (\ref{eq:prior}) is used, GP interpolation can be performed efficiently due to the Markovian property of the LTV-SDE given in (\ref{ltvsde}). State $x(\tau)$ at $\tau \in [t_i, t_{i+1}]$ is a function only of its neighboring states (\cite{gpmp2})
\begin{equation}
x(\tau) = \mu(\tau) + \Lambda(\tau)(x_i - \mu_i) + \Psi(\tau)(x_{i+1} - \mu_{i+1}),
\label{eq:intp1}
\end{equation}
\begin{equation}
\Lambda(\tau) = \Phi (\tau, t_i) - \Psi(\tau)\Phi(t_{i+1}, t_i),
\end{equation}
\begin{equation}
\Psi(\tau) = Q_{i, \tau} \Phi(t_{i+1}, \tau)^T Q_{i, i+1}^{-1},
\end{equation}
 where
 \begin{equation}
Q_{a, b} = \int_{t_a}^{t_b} \Phi(b, s) F(s) Q_c F(s)^T \Phi(b, s)^T ds.
\label{eq:Qab}
\end{equation}
The fact that any state $x(\tau)$ can be computed in $\mathcal{O}(1)$ complexity can be exploited for efficient computation of obstacle avoidance costs, as explained in Section \ref{subsec:multiag}.

Another major benefit arising from the aforementioned Markovian property of the LTV-SDE in (\ref{ltvsde}) is the fact that the inverse kernel matrix $K^{-1}$ of this prior is exactly sparse block tridiagonal (\cite{barfoot2014batch})
\begin{equation}
K^{-1} = A^{-T} Q^{-1} A^{-1}
\end{equation}
where
\begin{equation}
A^{-1} =
\begin{bmatrix}
1 & 0 & ... & 0 & 0 \\
-\Phi(t_1, t_0) & 1 & ... & 0 & 0 \\
0 & -\Phi(t_2, t_1) & \ddots & \vdots & \vdots \\
\vdots & \vdots & \ddots & 1 & 0 \\
0 & 0 & ... & -\Phi(t_N, t_{N-1}) & 1
\end{bmatrix}
\end{equation}
and
\begin{equation}
Q^{-1} = \text{diag} (K_0^{-1}, Q_{0,1}^{-1}, ... , Q_{N-1, N}^{-1})
\end{equation}
with $Q_{a,b}$ given in (\ref{eq:Qab}), the trajectory going from $t_0$ to $t_N$, and $K_0$ being the initial covariance. As it will be shown in Section \ref{subsec:multiag}, this kernel allows for fast, structure-exploiting inference.

\section{Multi-agent trajectory optimization as probabilistic inference}
\label{subsec:multiag}
% % To obtain the MAP trajectory by solving the aforementioned optimization problem with iterative approaches, such as Gauss-Newton or Levenberg-Marquardt, we need a linearized approximation of Eq. \ref{MAP2}.
% % (( more text will folow ))
% %
% % Since the system in equation(referenca) is linear and sparse, MAP trajectory optimization can be represented as the problem of inference on a factor graph, which allows exploiting sparsity for finding the solution efficiently. Posterior distribution given in Eq. \eqref{posterior}  can be factorized into GP, collision, manipulability and interpolation factors similar to (referenca):
% % \begin{multline}
% % P(x|e) \propto \\ \prod_{t_i} f^{gp} (x_i, x_{i+1}) f_{i}^{obs} (x_i)  f_{i}^{\mu}(x_i)
% %  \prod_{\tau = 1}^{n_p} f_{i, \tau}^{intp}(x_i, x_{i+1})
% % \end{multline}
% %
In this section we formulate the multi-agent trajectory optimization problem as probabilistic inference. The presented formulation is predicated on the work of \cite{gpmp2} and it represents the extension of their trajectory optimization method to multi-robot systems.

To formulate the trajectory optimization problem as probabilistic inference, we seek to find a trajectory parametrized by $x$ given desired events $e$.  The posterior density of $x$ given events $e$ can be computed via Bayes' rule from a prior and likelihood
\begin{align}\label{eq:posterior}
p(x | e) = p(x) p(e | x) / p(e) \propto p(x) p(e|x)
\end{align}
where $p(x)$ represents the prior on $x$ which encourages smoothness of the trajectory, while $p(e| x)$ represents the probability of the desired events occuring given $x$. The optimal trajectory $x$ is found by maximizing the posterior $p(x | e)$, using the \textit{maximum a posteriori} (MAP) estimator
\begin{align}\label{eq:MAP}
x^* = \underset{x}{\arg\max} \, p(x) p(e | x),
\end{align}
where
\begin{equation}
 p(e | x) = \prod_i p(e | x_i)
\end{equation}
with $x_i$ being the configuration at discrete time instant $t_i$.
The conditional distributions $p(e | x_i)$ specify the likelihood of desired events occuring at the configuration $x_i$
\begin{align}\label{eq:conditional1}
L (x_i | e)  \propto p(e | x_i).
\end{align}

In motion planning context, desired event is avoidance of collisions and therefore conditional distribution $p(e | x_i)$ specifies the likelihood that the configuration $x_i$ is collision free
\begin{align}\label{eq:conditional1}
L_{obs} (x_i | c_i = 0)  \propto p(c_i = 0| x_i).
\end{align}

In our case, for each agent there are two possible types of collision; collision with a static obstacle in the environment and collision with another agent. Since the aforementioned types of collision are independent events, the likelihood of trajectory $x$ being free of collisions can therefore be considered as the product of two likelihoods:
\begin{equation}
L_{obs} (x | c=0) =  L_{stat} (x | c_{obs}=0) L_{mul} (x | c_{mul}=0),
\end{equation}
where the first the term specifies the probability of being clear of collisions with static obstacles and the second term specifies the probability of being clear of collisions with other agents.

For each agent $j$, the likelihood of being free of collision with static obstacles is defined as a distribution of the exponential family  \citep{gpmp2}:
\begin{align}
L_{stat} (x_j | c_{obs} = 0)  \propto \exp \{ -\frac{1}{2} \norm{ h(x_j) }^2_{{\Sigma_{obs}}} \},
\label{eq:obslike}
\end{align}
where $h(x)$ is a vector-valued obstacle cost function and $\Sigma_{obs}$ a diagonal matrix and the hyperparameter of the distribution. In the same manner, we define the likelihood of being collision-free with other agents as a distribution of the exponential family that is a product of probabilities of being free of collision with every other agent
\begin{equation}
L_{mul} (x_{j} | c_{mul} = 0) \propto  \prod_{\substack{j^{\prime}=1 \\ j^{\prime} \neq j}}^{n_{ag}}  \exp \{ -\frac{1}{2} \norm{ g(x_j, x_{j^{\prime}}) }^2_{{\Sigma_{mul}}}    \},
\label{eq:multlike}
\end{equation}
where $g(x_j, x_{j^{\prime}})$ is a vector-valued function that defines the cost of two agents $j$ and $j^{\prime}$ being close to each other, $n_{ag}$ is number of agents and $\Sigma_{mul}$ is a hyperparameter of the distribution.

Combining Eq. \eqref{eq:obslike} and Eq. \eqref{eq:multlike}, the total likelihood of agent $j$ being free of collision is obtained
\begin{multline}
L_{obs} (x_{j} | c = 0) \propto \\  \exp \{ - \frac{1}{2}  \norm{ h (x_{j})}_{\Sigma_{obs}}^2 \}  \prod_{\substack{j^{\prime}=1 \\ j^{\prime} \neq j}}^{n_{ag}}  \exp \{ - \frac{1}{2}  \norm{ g (x_{j}, x_{j^{\prime}} )}_{\Sigma_{{mul}}}^2 \}.
\label{eq:totallike}
\end{multline}

Deriving the MAP trajectory from Eq. \eqref{eq:MAP}, Eq. \eqref{eq:prior} and Eq. \eqref{eq:totallike} in a similar manner to \cite{gpmp2}, our \textit{maximum a posteriori} trajectory for each agent is
\begin{multline}
x_j^* = \underset{x_j}{\arg\min} \, \{ \frac{1}{2} \norm{x_j - \mu_j}_{K_j}^2 + \\ \frac{1}{2} \norm{ h(x_j)}^2_{\Sigma_{obs}} + \frac{1}{2} \sum_{\substack{j^{\prime}=1 \\ j^{\prime} \neq j}}^{n_{ag}}    \norm{ g (x_{j}, x_{j^{\prime}} )}_{\Sigma_{{mul}}}^2 \} .
\label{eq:MAP2}
\end{multline}
This is a nonlinear least squares problem that can be solved with iterative approaches, such as Gauss-Newton or Levenberg-Marquardt.

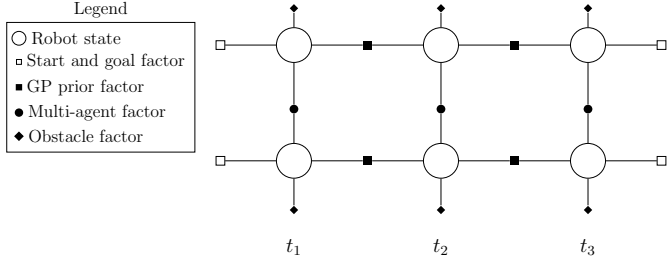
\begin{figure}[!t]
\centering
\vfill
\begin{minipage}[t]{.14\textwidth}
\centering
\vspace{4pt}
\resizebox {1\textwidth}{!}{\begin{tikzpicture}[baseline,
  pose/.style={shape=circle, draw=black, fill=white, line width=0.5, scale=1},
  fix/.style={shape=rectangle, draw=black, fill=white, line width=0.5, scale=0.5},
  gpprior/.style={shape=rectangle, draw=black, fill=black, line width=1, scale=0.5},
  multi/.style={shape=circle, draw=black, fill=black, line width=1, scale=0.5},
  obstacle/.style={shape=diamond, draw=black, fill=black, line width=1, scale=0.4}
]
   % \tikzstyle{every node}=[font=\Large]

\matrix [draw] (m) {
  \node [pose,label=right:{\large Robot state} ] {}; \\
  \node [fix,label=right:{\large Start and goal  factor}] {}; \\
  \node [gpprior,label=right:{\large GP prior factor}] {}; \\
  \node [multi,label=right:{\large Multi-agent factor}] {}; \\
  \node [obstacle,label=right:{\large Obstacle factor}] {}; \\
};
\node[anchor=south] at (m.north) {\large Legend};

\end{tikzpicture}  %
}
\end{minipage}
\begin{minipage}[t]{.34\textwidth}
\centering
\vspace{0pt}
\resizebox {1\textwidth}{!}{\begin{tikzpicture}[baseline]

  % Layout the variables
  \matrix[row sep=0.3cm, column sep=1.0cm]
  { %
     & 
     \factor[diamond] {fo11} {} {} {}; &
     &
     \factor[diamond] {fo12} {} {} {}; &
     &
     \factor[diamond] {fo13} {} {} {}; 

  \\
  	\factor[fill=white,draw=black] {f10} {} {} {}; &
    \node[latent] (11) {} ; & %
    \factor[] {f11} {} {} {}; &
    \node[latent] (12) {} ; & %
    \factor[] {f12} {} {} {}; &
    \node[latent] (13) {} ; & %
    \factor[fill=white,draw=black] {f13} {} {} {}; 
    \\
    \\
     & 
     \factor[circle] {fm1} {} {} {}; &
     &
     \factor[circle] {fm2} {} {} {}; &
     &
     \factor[circle] {fm3} {} {} {}; &

	\\
    \\
    \factor[fill=white,draw=black] {f20} {} {} {}; &
    \node[latent] (21) {} ; & %
    \factor[] {f21} {} {} {}; &
    \node[latent] (22) {} ; & %
    \factor[] {f22} {} {} {}; &
    \node[latent] (23) {} ; & %
    \factor[fill=white,draw=black] {f23} {} {} {}; 
    \\
      & 
     \factor[diamond] {fo21} {} {} {}; &
     &
     \factor[diamond] {fo22} {} {} {}; &
     &
     \factor[diamond] {fo23} {} {} {}; &
	\\
		&
	\node[latent, fill=white,draw=white] (t1) {\large $t_1$} ;  &
	&
	\node[latent, fill=white,draw=white] (t2) {\large $t_2$} ;  &
	&
	\node[latent, fill=white,draw=white] (t3) {\large $t_3$} ;
	\\
  };
  
%start & goal unary factor edges
 \edge[-] {f10} {11};
 \edge[-] {f13} {13};
 \edge[-] {f20} {21};
 \edge[-] {f23} {23};
 
% obstacle unary factor edges 
 \edge[-] {fo11} {11};
 \edge[-] {fo12} {12};
 \edge[-] {fo13} {13};
 \edge[-] {fo21} {21};
 \edge[-] {fo22} {22};
 \edge[-] {fo23} {23};

% vertical factor edges
\factoredge[-] {11} {fm1} {21} ;
\factoredge[-] {12} {fm2} {22} ;
\factoredge[-] {13} {fm3} {23} ;

% horizontal factor edges
\factoredge[-] {11} {f11} {12} ;
\factoredge[-] {12} {f12} {13} ;
\factoredge[-] {21} {f21} {22} ;
\factoredge[-] {22} {f22} {23} ;

\end{tikzpicture}  %
}  %%graf4.tikz
\end{minipage}
\label{fig:factorgraph}
\caption{A simple illustration of the factor graph describing an example multi-robot trajectory optimization problem. GP interpolation factors present between states are omitted for clarity.}
\end{figure}

To obtain the MAP trajectory by solving the aforementioned optimization problem with iterative approaches, we need a linearized approximation of (\ref{eq:MAP2}). Converting the nonlinear least squares problem to a linear problem around operating point $\bar{x}_j$, the following expression for the optimal perturbation $\delta x_j^*$ is obtained
\begin{multline}
\delta x_j^* =  \underset{\delta x_j}{\arg\min} \Big \{ \frac{1}{2} \norm{\bar{x}_j + \delta x_j - \mu_j}_{K_{j}}^2 + \\ \frac{1}{2} \norm{h(\bar{x}_j) + H_j \delta x_j}_{\Sigma_{obs}}^2 + \\  \frac{1}{2} \sum_{\substack{j^{\prime}=1 \\ j^{\prime} \neq j}}^{n_{ag}}  \norm{g (\bar x_{j}, x_{j^{\prime}}) + G_j\delta x_j}_{\Sigma_{mul}}^2
 \Big  \} ,
 \label{eq:MAP3}
\end{multline}
where $H_{j}$ is the Jacobian matrix of $h(x_j)$
\begin{equation}
H_{j} = \frac{\partial h}{\partial x_{j}} \bigg|_{\bar {x}_j},
\end{equation}
and $G_{j}$ is the partial derivative of $g(x_j, x_{j^{\prime}})$
\begin{equation}
G_{j} = \frac{\partial g}{\partial x_{j}} \bigg|_{(\bar{x}_j, x_{j^{\prime}})}.
\end{equation}
The optimal perturbation $\delta x_j^*$ is obtained by solving the following linear system
\begin{multline}
\bigg ( K_j^{-1} + H_j^{T} \Sigma_{obs}^{-1} H_j + \sum_{\substack{j^{\prime}=1 \\ j^{\prime} \neq j}}^{n_{ag}} G_j^{T} \Sigma_{mul}^{-1} G_j \bigg ) \delta x_j^* = \\ K_j^{-1} (\mu_j - \bar{x}_j) - H_j^{T} \Sigma_{obs}^{-1} h (\bar{x}_j) - \sum_{\substack{j^{\prime}=1 \\ j^{\prime} \neq j}}^{n_{ag}}  G_j^{T} \Sigma_{mul}^{-1}g(\bar{x}_j, x_{j^{\prime}}).
\label{eq:linearsystem}
\end{multline}
 This MAP trajectory optimization can be represented as inference on a factor graph (\cite{kschischang2001factor}). The fact that the system in (\ref{eq:linearsystem}) is linear and sparse can be exploited for finding the solution efficiently. In our case, the posterior distribution given in Eq. \eqref{eq:posterior}  can be factorized similarly to \cite{gpmp-ijrr}:
\begin{multline}
P(x|e) \propto  \prod_{t_i} \prod_{c_j} f_j^{gp} (x_i^j, x_{i+1}^j) f_{i,j}^{obs} (x_i^j)   \\ \prod_{c_{j^{\prime}}} f^{mul}_{i, j, j^{\prime}} (x_i^j, x_i^{j^{\prime}}) \prod_{\tau = 1}^{n_p} f_{i,j, j^{\prime}, \tau}^{intp}(x_i^j, x_{i+1}^j, x_i^{j^{\prime}}, x_{i+1}^{j^{\prime}}),
 \label{eq:factorgraph}
\end{multline}

where $f^{gp}$ represents factor corresponding to a GP prior, $f^{obs}$ represents the cost of collision with static obstacles, $f^{mul}$ represents the cost of collision with other agents and $f^{intp}$ represents collision cost calculated for $n_p$ interpolated states that can be obtained by using (\ref{eq:intp1}). The factor graph defined in (\ref{eq:factorgraph}) is depicted in Fig. \ref{fig:factorgraph} for a simple case of trajectory optimization problem with two agents and three states.

%\subsection{Simultaneous trajectory optimization}
%
%Each edge in the graph connects its two states via a GP prior %factor

\section{Implementation details} \label{subsec:impdet}
\subsection{GP prior}
In our implementation, for dynamics of our robot we use the double integrator linear system with white noise injected in acceleration, meaning that the trajectory is generated by the LTV-SDE in (\ref{ltvsde}) with
\begin{equation}
A = \begin{bmatrix}
0 & \text{I} \\
0 & 0
\end{bmatrix}, \, u(t) = 0, \, F(t) = \begin{bmatrix}
0 \\ \text{I}
\end{bmatrix}.
\end{equation}
This represents a constant velocity GP prior which is centered around a zero acceleration trajectory (\cite{barfoot2014batch}). Applying such a prior will minimize actuator velocity in the configuration space, thus minimizing the energy consumption and giving the physical meaning of smoothness.
Choosing the noise covariance $Q_c$ has an effect on smoothness, with smaller values penalizing deviation from the prior more.

\subsection{Avoidance of collision with static obstacles}
In (\ref{eq:obslike}) we defined the likelihood of one agent $j$ being free of collision with static obstacles which relies upon a vector-valued obstacle cost function $h(x_j)$. In our case, for the obstacle cost $h(x_j)$ we use the hinge loss function with a precomputed signed distance field similarly to \cite{gpmp2} and \cite{gpmp-ijrr}.
The pertaining hinge loss function is defined as:
\begin{equation}
h(x_j) = \begin{cases}
\centering
                   \varepsilon_{obs} - d_{s,j} &\mbox{ if } d_{s,j} \leq \varepsilon_{obs} \\
                \hfil  0  &\mbox{ if } d_{s,j} > \varepsilon_{obs}
                \end{cases},
                \label{eq:hinge1}
\end{equation}
where $d_{s,j}$ is the signed distance, and $\varepsilon_{obs}$ is the \textit{safety distance} indicating the boundary of the \textit{danger area} near obstacle surfaces (\cite{gpmp-ijrr}).

\subsection{Avoidance of collision with other agents}
In (\ref{eq:multlike}) we defined the likelihood of one agent $j$ being free of collision with other agent $j^{\prime}$. This likelihood relies on a vector-valued function $g(x_j, x_{j^{\prime}})$, which we defined as the hinge loss function:
\begin{equation}
g(x_j, x_{j^{\prime}}) = \begin{cases}
\centering
                   \mathcal{\varepsilon}_{mul} - d_{j, j^{\prime}} &\mbox{ if } d_{j, j^{\prime}} \leq \varepsilon_{mul} \\
                \hfil  0  &\mbox{ if } d_{j, j^{\prime}} > \varepsilon_{mul}
                \end{cases},
                \label{eq:hinge}
\end{equation}
where $d_{j, j^{\prime}}$ denotes $d(x_j, x_{j^{\prime}})$, the Euclidean distance between two agents $j$, and $j^{\prime}$ and  $\varepsilon_{mul}$ is a \textit{safety distance} indicating the boundary of the area inside which we anticipate a collision. If the parameter $\varepsilon_{mul}$ is set to zero, multi-agent obstacle cost would always be zero and our planner would work exactly the same as GPMP2 (\cite{gpmp2}) for each agent. Higher values of $\varepsilon_{mul}$ enable robots to anticipate each other's future trajectories and to adapt them accordingly in the next iteration of optimization, giving our algorithm a predictive property.

To obtain the MAP trajectory in (\ref{eq:MAP3}), the proposed method requires $G_j$, the partial derivative of the hinge loss function defined in (\ref{eq:hinge}). This partial derivative used in our implementation can be analytically obtained:
\begin{equation}
\frac{\partial g(x_j, x_{j^{\prime}})}{\partial x_{j}} = \begin{cases}
\centering
                \frac{x_{j^{\prime}} - x_j }{d(x_j, x_{j^{\prime}})}. &\mbox{ if } d_{j, j^{\prime}} < \varepsilon_{mul} \\
                \hfil 0.5 &\mbox{ if } d_{j, j^{\prime}} = \varepsilon_{mul} \\
                \hfil  0  &\mbox{ if } d_{j, j^{\prime}} > \varepsilon_{mul}
                \end{cases}.
                \label{eq:hinge}
\end{equation}

Parameters $\Sigma_{obs}$ and $\Sigma_{mul}$, needed to fully implement static and multi-agent obstacle likelihoods in (\ref{eq:obslike}) and (\ref{eq:multlike}) are defined by isotropic diagonal matrices $\Sigma_{obs} = \sigma_{obs}^2 I$ and $\Sigma_{mul} = \sigma_{mul}^2 I$, where $\sigma_{obs}$ and $\sigma_{mul}$ represent \textit{obstacle cost weights}. Reducing the value of $\sigma_{obs}$ causes the optimization to place more weight on avoiding collision with static obstacles, while reducing the value of $\sigma_{mul}$ causes the optimization to place more weight on avoiding collision with other agents.

\subsection{Software implementation}

In our experiments we use the GPMP2 C++ library (\cite{gpmp2}, \cite{dong2017sparse}), and its respective MATLAB toolbox, which is based on the GTSAM C++ library (\cite{dellaert2012factor}, \cite{dellaert2006square}). Experiments are performed on a system with a 3.8-GHz Intel Core i7-7700HQ processor and 16 GB of RAM.

\section{Experimental results}
\label{subsec:results}
\subsection{Formation control}
\label{subsec:formcontrol}
In this section we present the results of our method in a quantitative manner. We used the proposed approach to switch the positions of 2D holonomic circular robots with radius $r = 1$\,m inside a formation in an obstacle free simulation environment. We did this for every possible permutation of positions inside a formation for cases of three, four and five agents and compared the success rate and computation time of our approach to the GPMP2 framework (\cite{gpmp2}) where each agent's trajectory is planned individually at every time step. In three and five robot simulation the formation is a triangle, while in four robot simulation it's a square. That means that we tested and compared algorithms on the total of $150$ unique planning problems. When using the GPMP2 framework, for every agent the others were incorporated in its signed distance field (SDF), meaning that at every time step the SDF had to be changed and the trajectory had to be replanned.

All trajectories were initialized as a constant-velocity straight line trajectory in configuration space. Total time of execution for every case is set as $t_{total} = 10 $\,s and all trajectories were parametrized with $10$ equidistant support states such that $9$ points are interpolated between any two states ($91$ states effectively). The parameters used for GPMP2 are $\varepsilon_{obs} = 2$, $\Sigma_{obs} = 0.3$, while the parameters used for our approach are $\varepsilon_{mul} = 15$, $\Sigma_{mul} = 0.7$. It makes sense to use relatively small $\varepsilon_{obs}$ in comparison to $\varepsilon_{mul}$ since in GPMP2 the agents only react to each other locally, and in our approach it is desireable that every agent knows each other's position at all times in order to calculate the trajectories that avoid collisions. The success rates comparison is showed in Table \ref{tab:succ} and the computational times comparison is showed in Table \ref{tab:time}.

\begin{table}[!t]
\begin{center}
\caption{Comparison of success rates (percentages) for our method and replanning with GPMP2 for every possible permutation of positions inside formations of 3, 4 and 5 agents}\label{tb:successrate}
\begin{tabular}{ccc}
\hline
Number of robots & GPMP2 & MUL-GPMP  \\\hline
3 & 100 & 100  \\
4 & 87.5 & 100  \\
5 & 70.8 & 100  \\ \hline
\label{tab:succ}
\end{tabular}
\end{center}
\end{table}

\begin{table}[!t]
\begin{center}
\caption{Comparison of average execution times ($ms$) for our method and replanning with GPMP2 for every possible permutation of positions inside formations of 3, 4 and 5 agents}\label{tb:avgtime}
\begin{tabular}{ccc}
\hline
Number of robots & GPMP2 & MUL-GPMP  \\\hline
3 & 196 & 22  \\
4 & 264 & 37  \\
5 & 457 & 59  \\ \hline
\label{tab:time}
\end{tabular}
\end{center}
\end{table}

\begin{figure}[!t]
%\hspace{4pt}
\centering
\resizebox {0.48\textwidth}{!}{\input{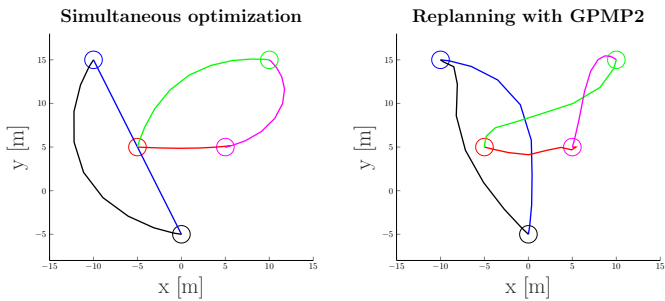}}
\caption{Example of switching formation with the proposed simultaenous optimization and with replanning with GPMP2. The proposed algorithm produces visibly smoother trajectories due to its predictive nature.}
\label{fig:exp2}
\end{figure}

 Since computing the SDF and replanning the trajectory is relatively computationally demanding in comparison to computing the \textit{multi-agent} factor in (\ref{eq:multlike}), it is not surprising that our approach achieves significantly faster computational times. Incorporating current positions of other agents inside an agent's SDF means that the GPMP2 framework, unlike the proposed method, can only react to the changes in the environment and not anticipate them. Thus the better success rates of our approach were also expected. Our method was able to successfully solve all of the given 150 motion planning problems, while the GPMP2 failed in total 38 cases. Due to the predictive nature of the proposed approach, the generated trajectories are also visibly smoother which is demonstrated in Fig.~\ref{fig:exp2} for one example of five robot formation change. Note that for the case of replanning with GPMP2 better results could be achieved by parameter tuning, possibly using a grid search, but that is relatively computationally demanding and possibly unattainable in the real world circumstances.

\subsection{Complex static environment}
\label{subsec:labirint}
This experiment shows qualitatively the capability of our approach to generate trajectories that can ensure successful simultaneous motion of two agents in complex static environments.
Our goal was to switch positions of two planar holonomic robots that start in two rooms which are separated by a narrow hallway. The radius of each robot is $r = 1$\,m, and the width of the hallway is $w = 3.6$\,m, which is smaller than two robots diameters combined, meaning that robots cannot pass each other in the hallway. Since the environment is complex, if the initial trajectory of each robot is set as a straight line, the optimization gets stuck in the local minimum and the collisions are not avoided. To solve that problem, we set the initial trajectories as the paths obtained by $A^*$ algorithm with constant velocities. When computing the path with  the $A^*$ algorithm, we downsampled the grid representing the environment so that the initial path was obtained in acceptable time. Since the paths obtained by the $A^*$ algorithm are only used as priors in our method, possible loss of information about small sized obstacles when downsampling is not concerning.

\begin{figure*}
\hspace{4pt}
\resizebox {0.95\textwidth} {!}{\input{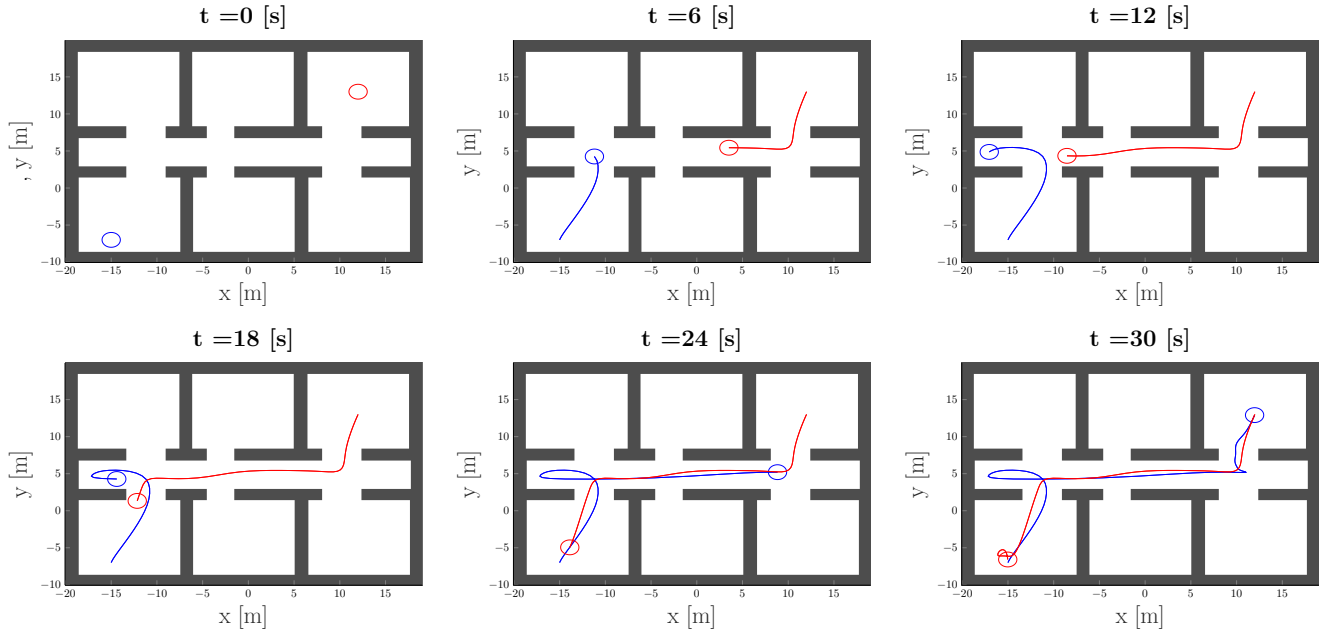}}
\caption{The example of trajectories planned with our approach for two agents in a complex static environment consisting of rooms with narrow passage between them. Robots anticipate each other's trajectories and one robot makes way for the other one. }
\label{fig:labirint}
\end{figure*}

The described setup represents a challenging task for existing motion planning algorithms, and most of them would result in redundant motion; robots would move towards each other and meet in the middle of the hallway, where one of them would have to turn back to make way for the other. In our case, however, due to the fact that the proposed approach works in a predictive manner, after exiting the room one robot turns away from its goal and waits for the other one to enter its destination room before proceeding to travel to its goal. The downside of the proposed method is the number of the optimization hyperparameters that need to be set, for example static and multi-agent obstacle cost factor covariances $\Sigma_{obs}$ and $\Sigma_{mul}$. In this specific example, the optimization hyperparameters $\Sigma_{obs}$, $\Sigma_{mul}$, $\varepsilon_{obs}$, $\varepsilon_{mul}$ were tuned via exhaustive grid search. The described environment and the result of the conducted experiment are shown in Fig. \ref{fig:labirint}.

%Note that in order to achieve this result,  which is computationally demanding, and . Reducing the number of

\section{Conclusion and future work}
\label{subsec:conclusion}
In this paper we have presented a fast trajectory optimization method for multi-agent motion planning. We considered each trajectory as a sample from a continuous time Gaussian process generated by linear time-varying stochastic differential equation driven by white noise. We formulated the multi-agent planning problem as probabilistic inference on a factor graph, and thus were able to exploit the structure of the mentioned GP to find the solution efficiently using numerical optimization. The proposed approach works in a predictive manner since each agent's trajectory is optimized simultaneously.  We tested our approach in simulation and compared it to planning each trajectory indvidually, demonstrating significant improvement in both success rate and computational efficiency.

In future work, it would be interesting to investigate how different priors affect the result of optimization. Another potentially interesting direction would be to explore the possibility of planning robot's trajectory in a dynamic environment. If the trajectory of a dynamic obstacle could be predicted, the avoidance of such obstacle would be achieved using the same concepts introduced in the proposed approach.
\bibliography{ifacconf}
\end{document}